\documentclass{article}

\usepackage[preprint]{neurips_2026}

\usepackage[utf8]{inputenc} % 允许 utf-8 输入
\usepackage[T1]{fontenc}    % 使用 8-bit T1 字体
\usepackage{hyperref}       % 超链接
\usepackage{url}            % 简单 URL 排版
\usepackage{booktabs}       % 专业质量的表格
\usepackage{amsfonts}       % 黑板数学符号
\usepackage{nicefrac}       % 紧凑的分数符号
\usepackage{microtype}      % 微观排版优化
\usepackage{xcolor}         % 颜色支持

\usepackage{graphicx}
\usepackage{multirow}
\usepackage{wrapfig}
\usepackage[most]{tcolorbox}
\usepackage{amssymb}
\usepackage{algorithm}
\usepackage{titletoc}

\title{Training with Harnesses: On-Policy Harness Self-Distillation for Complex Reasoning}

\author{%
  Zhengyang Zhao\thanks{Equal contribution.} \\
  Peking University \\
  \texttt{zhengyangzhao25@stu.pku.edu.cn} \\
  \And
  Lu Ma\footnotemark[1] \\
  Peking University \\
  \texttt{maluqaq@163.com} \\
  \And
  Wentao Zhang\thanks{Corresponding author.} \\
  Peking University \\
  \texttt{wentao.zhang@pku.edu.cn} \\
}

\begin{document}

\maketitle

% --- 正文部分 ---
% 确保你的项目目录下有 sections/ 文件夹及对应的 .tex 文件
\begin{abstract}
Inference-time harnesses substantially improve large language models on complex reasoning tasks. However, the intrinsic capabilities of the underlying model remain unchanged by the addition of these external workflows. To bridge this gap, we introduce \emph{On-Policy Harness Self-Distillation} (OPHSD), which employs the harness-augmented current model as a teacher for self-distillation, thereby introducing extra supervisory signals from the harness beyond training data.  OPHSD internalizes task-specific harness capabilities into the student model, yielding robust generalizability and strong standalone performance across diverse reasoning tasks. Evaluated across draft--verify harness for text classification and plan--solve for mathematical reasoning tasks, OPHSD consistently outperforms strong baselines (e.g., +10.83\% over OPSD on HMMT25). Our analysis further indicates that reattaching the harness during inference yields no additional benefits and can even degrade performance, suggesting that complex harnesses need not always be permanent fixtures; instead, they can serve as temporary training scaffolds whose benefits are permanently fed back into the base model. Our code and training data are available at \url{https://github.com/zzy1127/OPHSD-On-Policy-Harness-Self-Distillation}.
\end{abstract}
\section{Introduction}
\label{sec:introduction}

Large language models (LLMs) perform strongly on many reasoning, coding, and agentic tasks ~\cite{claude46,openai2025gpt54,gemini,team2026kimi,zeng2026glm, deepseekai2026deepseekv4, huang2026step}. However, when used without an inference-time harness, they can remain brittle on problems that require evidence accumulation, intermediate verification, or long-horizon problem solving. In these settings, a model may overlook relevant context, propagate early mistakes, or fail to revise an incorrect partial solution. To address this limitation, recent systems pair a base model with an inference-time harness, an external scaffold that structures the model's inference process by controlling how information is maintained, retrieved, transformed, and presented to the model throughout inference ~\cite{openclaw2026,anthropic_claude_code_docs,chen2026codexharness,hermesagent2026, chen2024magdistructureddistillationmultiagent}. Such harnesses often improve robustness in practice, but the gain comes from the external procedure rather than from the model alone.

This separation between model and harness creates both practical and scientific challenges. In practice, these systems add latency, token cost, and engineering complexity, while introducing new failure modes in retrieval, transformation, and control flow. Scientifically, they make it harder to tell what the model itself has learned. The harness may supply a useful procedure for a given input, but once the harness is removed, the underlying model need not retain that procedure. The overall system can improve even while the model itself remains brittle.

Existing post-training methods do not directly close this gap. Supervised fine-tuning (SFT) imitates static demonstrations, but does not teach the model to adapt its procedure. Reinforcement learning (RL) optimizes task-level rewards, but often provides sparse supervision that weakly identifies which procedural behaviors matter. On-policy distillation (OPD) ~\cite{lu2025onpolicydistillation,song2026survey, agarwal2024onpolicydistillationlanguagemodels}, which trains on the student's own trajectories under dense token-level supervision from a teacher, is therefore a natural vehicle for internalizing the behavior of an inference-time harness. This leaves a central question: can the step-by-step procedure induced by such a harness be absorbed into the model parameters?

We study this question through \emph{on-policy harness self-distillation} (OPHSD), a self-distillation method built around a task-specific inference harness. During training, OPHSD runs the model inside that harness, so its rollouts are generated under an enhanced inference procedure, such as retrieval-augmented reasoning for online text classification or plan--solve orchestration for mathematical reasoning. These harness-assisted rollouts then serve as targets for training the same model without the extra scaffolding, so the student learns from its own assisted behavior. Formally, OPHSD uses a reverse-KL objective to match these on-policy rollouts. Intuitively, this encourages the unassisted model to reproduce the behavior observed under the stronger procedure. At inference time, the harness is removed, so any resulting improvement reflects behavior the model has learned to produce on its own rather than continued external assistance.

Empirically, OPHSD improves task performance while reducing reliance on the inference-time harness. Across online text classification and mathematical reasoning, it outperforms strong baselines and achieves the best overall results on the evaluated benchmarks. On text classification, it achieves the best results on LawBench accusation classification and USPTO reaction-type prediction; on mathematical reasoning, it attains the highest average pass@8 across AIME24, AIME25, OlympiadBench, and HMMT25 (10.83\% improve over OPSD and 8.33\% over GRPO on HMMT25). We also find that reattaching the harness at inference time yields no further gains and, for text classification, can even reduce performance.

This paper makes three contributions. First, we introduce OPHSD, a self-distillation approach that uses a model's own harness-assisted rollouts to train the same model without the harness. Second, we show that OPHSD applies across two qualitatively different harnesses: draft--verify for online text classification and plan--solve for mathematical reasoning. Third, we provide empirical evidence that the distilled model can internalize useful aspects of the harness, making the harness unnecessary, and in some cases counterproductive, at test time. More broadly, our results suggest a broader role for harnesses in LLM systems: beyond serving as inference-time scaffolds, they can also act as temporary training-time structures whose procedural benefits are transferred back into the model.

\vspace{-2mm}
\section{Related Work}
\label{sec:related_works}

\subsection{On-Policy Distillation}
Knowledge distillation transfers the capabilities of a teacher model to a student model. Conventional distillation trains the student on teacher-generated responses or human demonstrations, which can create a mismatch between training and inference. On-policy distillation (OPD) ~\cite{song2026survey,lu2025onpolicydistillation} instead trains on trajectories sampled from the student’s current policy, with the teacher providing supervision on those student-generated rollouts. This allows the teacher to correct the states the student actually visits, reducing distribution mismatch during autoregressive generation. Existing OPD methods differ primarily in how they exploit teacher signals. Reward-based OPD formulates teacher–student divergence, such as reverse KL, as a policy-gradient reward ~\cite{lu2025onpolicydistillation,ko2026scaling,yang2026self,xiao2026mimo}. This paradigm naturally accommodates non-differentiable or trajectory-level feedback, but it can suffer from high variance and sensitivity to reward design. By contrast, Loss-based OPD directly restores differentiable token-level distillation losses in student-generated rollouts ~\cite{zhao2026selfdistilledreasoneronpolicyselfdistillation, shenfeld2026self, fu2026revisiting,deepseekai2026deepseekv4, he2026selfdistillationzeroselfrevisionturns}. It provides dense, lower-variance supervision, but requires access to teacher distributions and can be computationally expensive when the teacher is large. OPHSD is adjacent to OPD but differs from the standard teacher-scored setup. It also uses student-conditioned trajectories, but its supervision comes from traces generated by the base model inside an external harness, not from a separate teacher distribution over the same prefixes. Its goal is to internalize harness-induced response patterns so the distilled model can reproduce them without the harness at inference time.

\subsection{Harness Engineering}
A harness is the orchestration layer surrounding model or agent calls for a task family ~\cite{anthropic2025effectivecontextengineering}. It defines the control flow, intermediate artifacts, tool mediation, verification gates, stopping conditions, and state that persist across steps or delegated workers. Recent public engineering accounts identify harness engineering as a key driver of reliability and performance in agents ~\cite{chen2026codexharness,anthropic_claude_code_docs,openclaw2026,hermesagent2026}. In research, a distinct but related line of work treats the harness itself as an object of design and optimization. NLAHs externalizes harness behavior into editable natural-language specifications executed by a shared runtime, whereas Meta-Harness searches over harness implementations to improve downstream agent performance~\cite{lee2026metaharnessendtoendoptimizationmodel,pan2026natural}. Complementary efforts, including AutoHarness~\cite{lou2026autoharness}, explore harness synthesis and modular harness design. These works investigate how surrounding orchestration can be represented, modified, and optimized to improve agent behavior. OPHSD shares the premise that orchestration matters, but it optimizes a different artifact. Harness-engineering methods improve the external system that remains in the deployment loop. By contrast, OPHSD uses the harness to produce supervision for a model that is later deployed without that runtime. The relevant distinction, therefore, is what persists after optimization: the harness itself or the model trained on harness-produced outputs.

% =============================================================================
%  Section 3 -- Methodology.  English, single-column NeurIPS style.
%  Drop into the same NeurIPS template as Section 4.  Required additional
%  packages (in addition to booktabs + graphicx already used by Section 4):
%    \usepackage{amsmath, amssymb}
%    \usepackage{algorithm}
%    \usepackage{algpseudocode}
%  Architecture figure expected at:  figures/fig_arch_ophsd.pdf
% =============================================================================

\section{Methodology}
\label{sec:methodology}

\subsection{Preliminaries: On-Policy Distillation and Privileged Context}
\label{sec:method_background}

We build on on-policy distillation (OPD)~\citep{lu2025onpolicydistillation}, which trains a student policy $p_S$ on its own rollouts while a teacher policy $p_T$ supplies dense, token-level supervision. Given a problem distribution $\mathcal{S}$, the student first samples a trajectory $\hat y \sim p_S(\cdot \mid x)$; both policies then evaluate $\hat y$ token by token, and the student $\theta$ is updated to minimize
\begin{equation}
\label{eq:opd}
\mathcal{L}_{\mathrm{OPD}}(\theta)
\;=\;
\mathbb{E}_{x \sim \mathcal{S}}
\;\mathbb{E}_{\hat y \sim p_S(\cdot \mid x)}
\frac{1}{|\hat y|} \sum_{n=1}^{|\hat y|}
D\!\Big(p_T(\cdot \mid x, \hat y_{<n})
\;\Big\|\;
p_S(\cdot \mid x, \hat y_{<n})\Big),
\end{equation}
where $D$ is a token-level divergence.  Compared with
RL-style objectives, Eq.~\eqref{eq:opd} avoids credit assignment over
sparse rewards and is therefore considerably more sample-efficient.

\begin{table}[t]
\centering
\small
\begin{tabular}{@{}ll@{}}
\toprule
Method & Privileged context $X$ \\
\midrule
OPSD~\citep{zhao2026selfdistilledreasoneronpolicyselfdistillation},
SDFT~\citep{shenfeld2026selfdistillationenablescontinuallearning}
  & Verified reference solution $y^\star$ \\
SDPO~\citep{hübotter2026reinforcementlearningselfdistillation}
  & Environment feedback collected during rollout \\
CRISP~\citep{sang2026crispcompressedreasoningiterative}
  & Static ``be concise'' instruction prompt \\
OEL~\citep{ye2026onlineexperientiallearninglanguage}
  &  Experiential knowledge of the past trajectories \\
\bottomrule
\end{tabular}
\vspace{6pt}
\caption{Differences of privileged context of existing methods.}
\label{tab:selfdistill_privileged}
\end{table}
A common construction obtains $p_T$ by supplying the same model---typically parameterized as a frozen copy $\tilde{\theta}$ to maintain training stability---with a piece of \emph{privileged context} $X$ that is hidden from the student, i.e., $p_T(\cdot \mid x) \triangleq p_{\tilde{\theta}}(\cdot \mid x, X)$. As summarized in Table~\ref{tab:selfdistill_privileged}, existing self-distillation methods differ chiefly in their choice of $X$, ranging from a verified reference solution to environment feedback or a static instruction prompt.

\subsection{OPHSD: On-Policy Harness Self-Distillation}
\label{sec:method_ophsd}

Our goal is to internalize the procedural behavior of an inference-time harness into the model itself, so that the harness can be removed at test time. Viewed through the privileged-context lens of \S\ref{sec:method_background}, this goal exposes a structural mismatch with existing self-distillation methods: every method in Table~\ref{tab:selfdistill_privileged} treats $X$ as a \emph{static variable}---a reference solution, an environment trace, an instruction string---so the teacher's advantage over the student is informational rather than procedural. Such a static variable reveals what a good answer looks like, but not how it should be derived, leading to suboptimal performance. A harness, by contrast, is fundamentally a program that orchestrates retrievals, intermediate model calls, and verification steps to produce an answer. Internalizing a harness therefore requires generalizing the privileged context from a static variable into a harness-driven workflow.

We extend OPD with two changes that realize this generalization. First, inspired by the Learning Using Privileged Information (LUPI) paradigm \citep{JMLR:v16:vapnik15b}, we write $z(x)$ for the generalized privileged input, which encapsulates any oracle information accessible only during training (e.g., a reference solution or an asymmetric memory bank). Second, instead of passively appending $z(x)$ to the prompt, we introduce a programmatic \emph{harness} $H$ that actively orchestrates the reasoning process. Driven by the student parameters $\theta$, the harness processes $x$ and $z(x)$ into a structured terminal context $\mathcal{C}[H_\theta(x, z(x))]$. The teacher $\tilde{\theta}$ then evaluates this terminal context to provide token-level supervision:
\[
p_T(\cdot \mid x) \;\triangleq\; p_{\tilde{\theta}}(\cdot \mid \mathcal{C}[H_\theta(x, z(x))]).
\]

Formulating the teacher as a harness-driven process provides a unified and extensible framework. The harness abstraction covers a wide spectrum of complexities, from a trivial static wrapper (OPSD and SDFT) to sophisticated multi-agent reasoning programs. By distilling these harness-induced trajectories, the student internalizes capabilities far beyond what static data injection can provide. We term this framework \emph{On-Policy Harness Self-Distillation} (OPHSD).

\begin{figure*}[htbp]
    \centering
    \includegraphics[width=\linewidth]{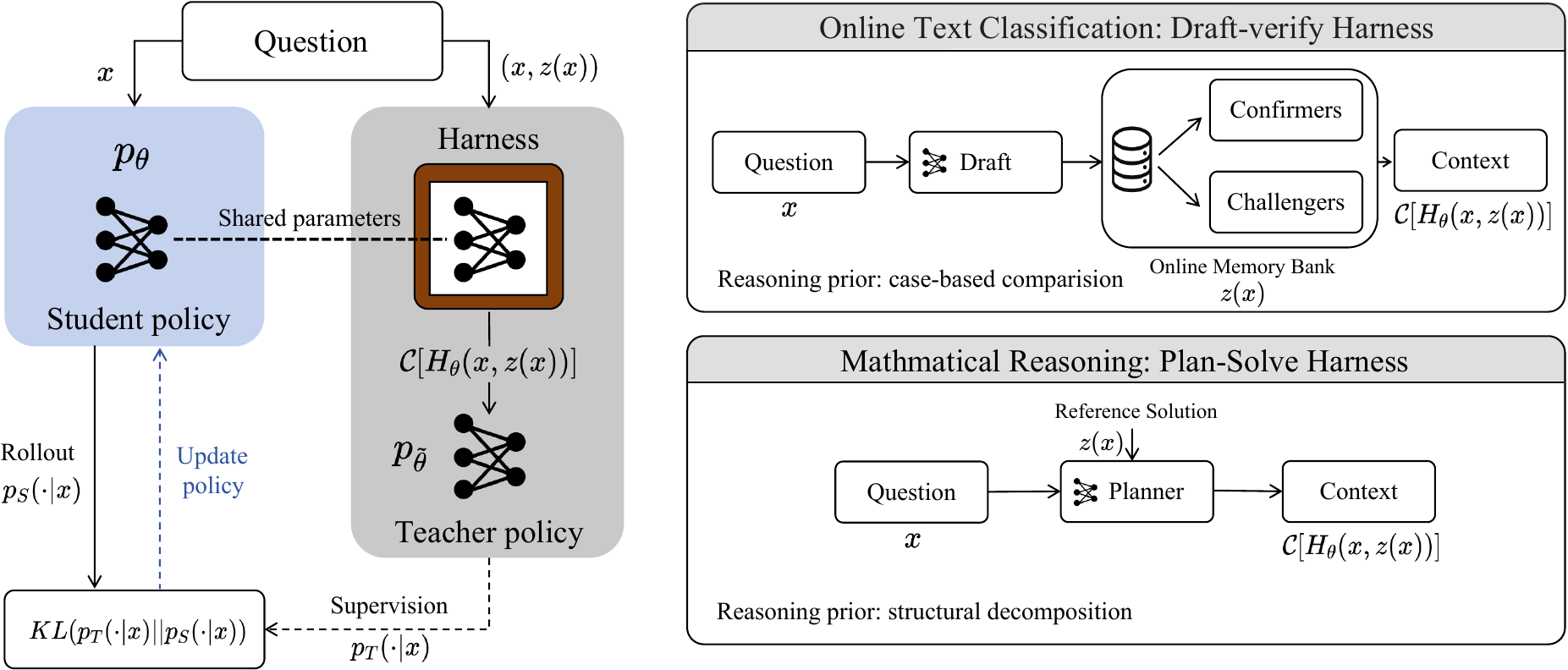}
    \caption{Overview of the OPHSD framework. 
    \textbf{Left:} The student policy simultaneously rollouts and interacts with the harness to generate trajectories, while the teacher generates supervisory signals based on harness trajectories and update the student by reverse KL.
    \textbf{Top Right:} Draft-verify Harness for text classification, using an online memory bank as the privileged input $z(x)$. 
    \textbf{Bottom Right:} Plan-Solve Harness for mathematical reasoning, using an oracle reference solution as $z(x)$.
    }
    \label{fig:ophsd_architecture}
\end{figure*}

% \begin{figure}[t]
% \centering
% \includegraphics[width=\linewidth]{figures/fig_arch_ophsd.pdf}
% \caption{\textbf{OPHSD architecture.}  The same parameters $\theta$ are
% shared by the student policy $p_S$ (single forward pass on $x$) and the
% teacher policy $p_T = H_\theta$, which executes a deterministic harness
% $H$ that issues multiple calls to $p_\theta$, optionally interacts with
% external tools, and emits a final response.  At each training step the
% student samples one direct rollout $\hat y \sim p_S(\cdot|x)$, both
% policies score $\hat y$ token by token, and gradients flow only through
% the student copy of $\theta$ (red arrow).  The harness's internal calls
% are treated as a stop-gradient target.}
% \label{fig:arch}
% \end{figure}
\paragraph{Harnesses as deterministic LLM programs.}
We follow the definition of \cite{lee2026metaharnessendtoendoptimizationmodel} and model a harness $H$ as a deterministic, stateful program that wraps a language model with parameters $\theta$. Concretely, $H$ maintains an internal state $s_t$, repeatedly issues queries to $p_\theta$ (and optionally to a fixed set of external tools), updates its state from the responses, and after at most $T$ such interactions, emits a final response $y$. By aggregating over these internal interaction trajectories, $H$ induces a conditional distribution:
\begin{equation}
\label{eq:harness_marginal}
H_\theta(y \mid x)
\;=\;
\sum_{s_{1:T}, \, c_{1:T}}
\Big[\prod_{t=1}^{T}
p_\theta(c_t \mid s_{t-1}, x)\,
\tau(s_t \mid s_{t-1}, c_t)\Big]
\;\delta\!\big(y = \pi(s_T)\big),
\end{equation}
where $c_t$ is the $t$-th model call, $\tau$ is the deterministic state transition function, and $\pi$ is a deterministic read-out that extracts the final answer from the terminal state. 

When $H$ additionally consumes a privileged input $z(x)$, the same construction defines $H_\theta(y \mid x, z(x))$ by injecting $z(x)$ into the harness's initial state, which enables the variant harness to provide students with stronger supervisory signals during training compared to when it is used solely for inference.

\paragraph{Training objective.}
During training, for each problem $x$, the student draws a single direct rollout $\hat y \sim p_\theta(\cdot \mid x)$. In parallel, we orchestrate a complete reasoning trajectory via the harness $H_\theta(x, z(x))$. Let $\mathcal{C}[H_\theta(x, z(x))]$ denote the terminal context produced by the dynamic harness. The student is then updated to minimize the reverse KL divergence against the target distribution computed by the static base model conditioned on this dynamic context:
\begin{equation}
\label{eq:ophsd}
\mathcal{L}_{\mathrm{OPHSD}}(\theta)
\;=\;
\mathbb{E}_{x \sim \mathcal{S}}
\;\mathbb{E}_{\hat y \sim p_\theta(\cdot \mid x)}
\frac{1}{|\hat y|} \sum_{n=1}^{|\hat y|}
\mathrm{KL}\!\Bigg(
\underbrace{p_{\tilde{\theta}}\big(\cdot \mid \mathcal{C}[H_\theta(x, z(x))],\, \hat y_{<n}\big)}_{p_T(\cdot \mid x,\, \hat y_{<n})}
\;\Bigg\|\;
\underbrace{p_\theta\big(\cdot \mid x,\, \hat y_{<n}\big)}_{p_S(\cdot \mid x,\, \hat y_{<n})}
\Bigg).
\end{equation}
where we treat $\mathcal{C}[H_\theta(x, z(x))]$ as a stop-gradient target. By decoupling the harness orchestration (driven by $\theta$) from the logit supervision (anchored by $\tilde{\theta}$), Eq.~\eqref{eq:ophsd} ensures that while the quality of the reasoning trajectory evolves alongside the model's growing capabilities, the supervisory signal remains grounded in the stable prior of the base model.

Moreover, because the student must reproduce this harness-induced behavior from the bare input $x$ alone, Eq.~\eqref{eq:ophsd} establishes a natural boundary on distillability. While structural reasoning priors (e.g., multi-step decomposition, self-verification) can be permanently internalized into the student's weights, genuine real-time external access (e.g., real-time tool interaction) cannot. OPHSD is therefore a recipe for internalizing scaffolded reasoning, not fabricating external access (further discussion in Appendix \ref{app:discussion}). To empirically demonstrate the generalizability of OPHSD, we next instantiate this framework across two distinct domains guided by this theoretical boundary.

\subsection{Harness Instantiations}
\label{sec:method_harnesses}

To validate OPHSD's generalizability, we apply it to two distinct reasoning modalities using off-the-shelf harnesses. We describe the cognitive intuition behind each selected harness below (full details in Appendix~\ref{app:harness}). 

\paragraph{Draft--verify for online text classification.}
For domain-specific text classification (e.g., legal charge prediction), the core challenge lies in distinguishing subtle boundaries among hundreds of confusable categories. The natural reasoning prior here is case-based comparison. Accordingly, the privileged input is an \emph{online memory bank} $z(x) = \mathcal{M}_{<x} = \{(x_i, y_i)\}_{i<t(x)}$ that collects all labelled precedents streamed in before $x$. We adopt the \emph{draft--verify} harness from \cite{lee2026metaharnessendtoendoptimizationmodel}, which queries $p_\theta$ twice on each input. In the \emph{draft} step, the harness retrieves the top-$k$ neighbours $\mathcal{N}_d(x) \subset \mathcal{M}_{<x}$ via sentence-embedding similarity, incorporates them as in-context demonstrations, and generates a draft prediction $\hat y_d \sim p_\theta(\cdot \mid x, \mathcal{N}_d(x))$. In the \emph{verify} step, $\hat y_d$ is used to re-query the bank for $k_+$ \emph{confirmers} $\mathcal{N}_+(x)$ (neighbours with the same label) and $k_-$ \emph{challengers} $\mathcal{N}_-(x)$ (neighbours with different labels). The harness then assembles a comprehensive prompt containing $x$, $\hat y_d$, and both retrieval sets to emit the final answer $y \sim p_\theta(\cdot \mid x, \hat y_d, \mathcal{N}_+(x), \mathcal{N}_-(x))$. Consequently, the terminal context consumed by the teacher in Eq.~\eqref{eq:ophsd} is $\mathcal{C}[H_\theta(x, z(x))] = (x, \hat y_d, \mathcal{N}_+(x), \mathcal{N}_-(x))$. At harness baseline evaluation, the memory bank is reset and re-populated exclusively from the test stream, ensuring no train/test leakage.

\paragraph{Plan--solve for mathematical reasoning.}
For mathematical reasoning, the primary challenge is bridging a concise problem statement to a highly specific answer via a long-horizon derivation. We formulate the reasoning prior here as structural decomposition. Using the reference solution as privileged input ($z(x) = y^\star$), the harness orchestrates a collaboration between a \emph{planner} and a \emph{solver} driven by the base model $p_\theta$. First, acting as the planner, the model leverages $(x, y^\star)$ to distill the oracle solution into a strategic sketch $s \sim p_\theta(\cdot \mid x, y^\star)$---a concise outline of key reasoning moves. Next, the solver takes $(x, s)$ to execute the complete algebraic derivation $y \sim p_\theta(\cdot \mid x, s)$. Unlike OPSD, which presents the oracle solution $y^\star$ directly verbatim, this harness transmits this solution through a mediating planner, utilizing this collaborative routing to force the student model to internalize the intrinsic structure of the derivation process. The terminal context consumed in Eq.~\eqref{eq:ophsd} is therefore $\mathcal{C}[H_\theta(x, z(x))] = (x, s)$, so the teacher signal the student matches is that of the solver—which has already observed the plan—rather than the $y^\star$ only. We also remove $y^\star$ at harness baseline evaluation, allowing the model to autonomously execute both roles from $x$ alone.
\section{Experiments}
\label{sec:exp}

We conduct experiments by the harnesses mentioned above on two tasks: online text classification and mathematical reasoning.

\subsection{Experimental Setup}
\label{sec:setup}

\paragraph{Model and datasets.}
All experiments start from Qwen3-8B\cite{yang2025qwen3technicalreport}.  For online text classification, we conduct two independent training runs (sampling 10k training examples each) to evaluate the following tasks:
(1) Legal Charge Prediction: We train on the CAIL-2018~\cite{xiao2018cail2018largescalelegaldataset} dataset and evaluate on LawBench~\cite{fei2024lawbench}, reporting the F1 score for predicting criminal charges from case descriptions (215 classes).
(2) Chemical Reaction Prediction: We train on the USPTO-50k~\cite{schneider2016s} dataset and evaluate on the USPTO test set, reporting accuracy for predicting reaction types from chemical equations (10 classes).
All training samples pass a strict contamination filter against both test sets.  For
mathematical reasoning, we sample $10$k problems from DeepMath\cite{he2025deepmath} as
training data and evaluate on four competition-level benchmarks:
AIME24, AIME25, OlympiadBench\cite{he2024olympiadbench} and HMMT25. We use 10\% data of OlympiadBench report the average performance of 4 runs.

\paragraph{Baselines.}
We compare four methods:
\textbf{GRPO}~\citep{shao2024deepseekmathpushinglimitsmathematical}
estimates a group-relative advantage over student rollouts and updates
the sampling policy with the resulting policy gradient.
\textbf{OPSD}~\citep{zhao2026selfdistilledreasoneronpolicyselfdistillation}
performs reverse-KL self-distillation where the teacher prefix is the
verified ground-truth label or solution. 
\textbf{CRISP}~\cite{sang2026crispcompressedreasoningiterative} teaches models to reason more concisely by distilling their own concise behavior and thereby enhancing accuracy (only for math tasks).
\textbf{OPHSD} (ours) dynamically constructs the teacher by wrapping the student model itself within a programmatic reasoning harness. We instantiate this framework using the draft--verify and plan--solve harnesses for the text and math domains, respectively. Following OPSD, we set $\tilde{\theta}$ to the initial policy to achieve stable optimization.  All methods
share the same training framework verl\cite{Sheng_2025} with
learning rate $1\!\times\!10^{-6}$, batch size $64$ and maximum
generation length $8192$ tokens. See Appendix \ref{app:baselines} for more details.

%-------------------------------------------------------------------
\subsection{Online Text Classification Results}
\label{sec:text}

\begin{table}[h]
  \centering
  \begin{minipage}[t]{0.38\linewidth}
    \vspace{0pt}\centering
    \resizebox{\linewidth}{!}{%
    \begin{tabular}{lcc}
      \toprule
      Method & LawBench (F1) & USPTO (Acc) \\
      \midrule
      Base           & 55.29 & 30.07 \\
      Harness & 60.22 & 79.02 \\
      \midrule
      GRPO                        & 62.44 & 90.01 \\
      OPSD                        & 64.25 & 88.01 \\
      OPHSD + Harness             & 68.41 & 83.62 \\
      \textbf{OPHSD}              & \textbf{69.51} & \textbf{90.81} \\
      \bottomrule
    \end{tabular}}
    \vspace{6pt}
    \caption{Online text classification results (\%).}
    \label{tab:text-peak}
  \end{minipage}\hfill
  \begin{minipage}[t]{0.60\linewidth}
    \vspace{0pt}\centering
    \resizebox{\linewidth}{!}{%
    \begin{tabular}{lccccc}
      \toprule
      Method & AIME24 & AIME25 & OlympiadBench & HMMT25 & Avg \\
      \midrule
      Base           & 23.33 & 16.67 & 55.88 &  6.67 & 25.64 \\
      Harness        & 40.00 & 40.00 & 64.71 & 29.17 & 43.47 \\
      \bottomrule
    \end{tabular}}
    \vspace{6pt}
    \caption{Performance of Qwen3-8B with and without
    the plan--solve harness (\%) on mathematical reasoning benchmarks. The harness leads the same model
    by $17.83\%$ on average.}
    \label{tab:math-step0}
  \end{minipage}
\end{table}

\begin{figure}[t]
  \centering
  \includegraphics[width=0.92\linewidth]{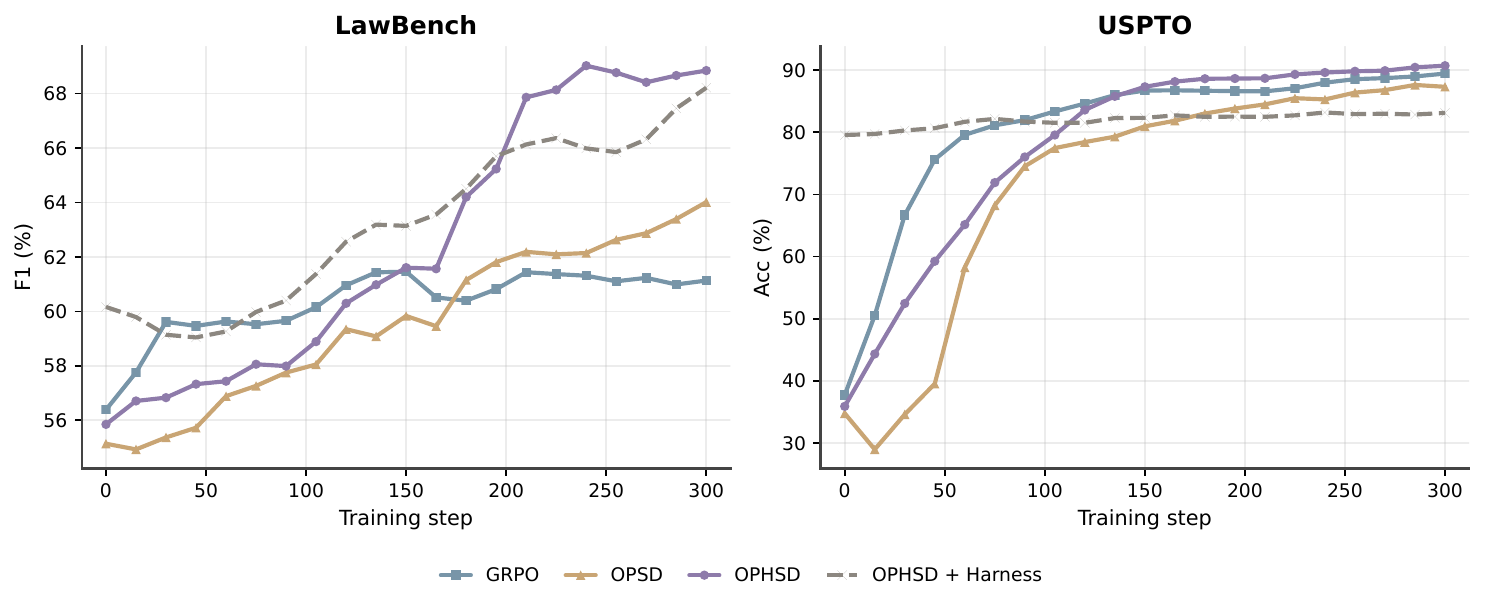}
  \caption{Evaluation performance on online text classification across the training stage. We test OPHSD + Harness on the memory bank built online from the test data.}
  \label{fig:text-train}
\end{figure}

We first validate the effectiveness of the harness, as shown in Table \ref{tab:text-peak}, on the base model, the draft--verify harness raises
LawBench F1 score by 4.93\% and USPTO accuracy by 48.95\%, indicating that the harness provides a sizeable signal that
distillation can exploit.  After training, OPHSD attains the highest
score on both benchmarks (69.51\% / 90.81\%), exceeding GRPO by
7.07\% / 0.80\% and OPSD by 5.26\% / 2.80\%. Notably, Figure \ref{fig:text-train} shows that as training progressed, OPHSD's reasoning capabilities gradually surpassed those of its harness. After distillation, OPHSD+Harness reduces its score by
$1.10\%$ on LawBench and $7.19\%$ on USPTO, which indicates that the harness no longer
yields a useful inference-time signal once it has been internalized.

\begin{wraptable}{r}{0.55\linewidth}
  \centering
  \small
  \setlength{\tabcolsep}{4pt}
  \begin{tabular}{llrrr}
    \toprule
                                & Method & early & mid & late \\
    \midrule
    \multirow{3}{*}{LawBench}   & GRPO            & 6.3   & 2.5    & 5.0   \\
                                & OPSD            & 5.0   & 5.0    & 2.5   \\
                                & \textbf{OPHSD}  & \textbf{7.5} & \textbf{100.0} & \textbf{97.5} \\
    \midrule
    \multirow{3}{*}{USPTO}      & GRPO            & 5.0   & 5.0    & 2.5   \\
                                & OPSD            & 5.0   & 2.5    & 7.5   \\
                                & \textbf{OPHSD}  & \textbf{7.5} & \textbf{75.0}  & \textbf{90.0}  \\
    \bottomrule
  \end{tabular}
  \caption{Cite rate across the training stage (\%).}
  \label{tab:text-cite}
\end{wraptable}

%-------------------------------------------------------------------
\subsection{Mathematical Reasoning Results}
\label{sec:math}

\begin{figure}[!h]
  \centering
  \includegraphics[width=\linewidth]{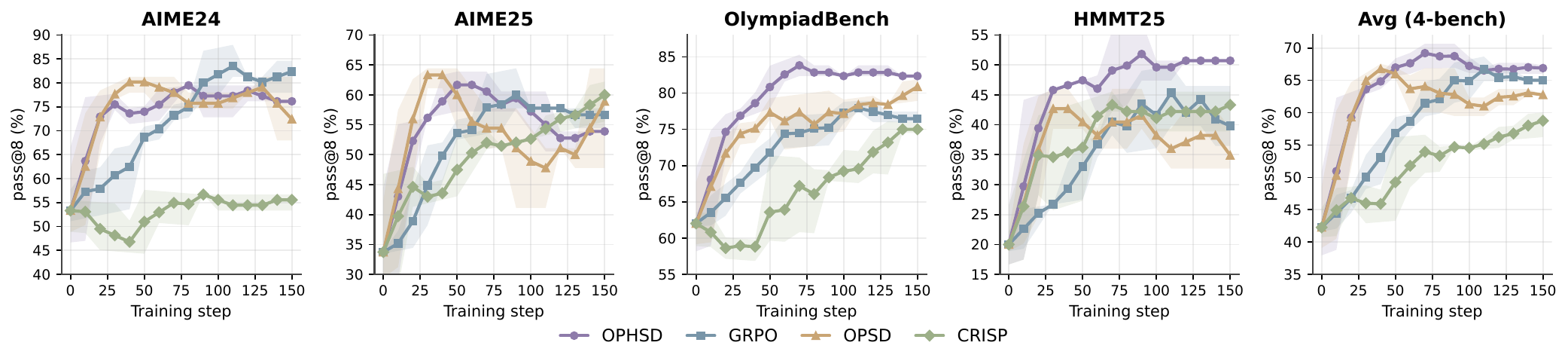}
  \caption{Evaluation of pass@8 performance across the training stage on
  four math benchmarks (\%). Shaded regions denote the fluctuation range.}
  \label{fig:math-train}
\end{figure}

\begin{table}[!h]
  \centering
  \small
  \setlength{\tabcolsep}{6pt}
  \begin{tabular}{lccccc}
    \toprule
    Method & AIME24 & AIME25 & OlympiadBench & HMMT25 & Avg \\
    \midrule
    GRPO              & \textbf{83.33} & 60.00 & 77.94 & 45.00 & 66.57 \\
    OPSD              & 80.00 & \textbf{63.33} & 80.88 & 42.50 & 66.68 \\
    CRISP             & 56.67 & 60.00 & 75.00 & 43.33 & 58.75 \\
    OPHSD + Harness   & 80.00 & \textbf{63.33} & 79.85 & \textbf{53.33} & 69.13 \\
    \textbf{OPHSD}    & 79.17 & 61.67 & \textbf{83.82} & \textbf{53.33} & \textbf{69.50} \\
    \bottomrule
  \end{tabular}
  \vspace{6pt}
  \caption{Mathematical reasoning results (\%).  We report pass@8 accuracy of the four benchmarks. OPHSD + Harness
  test the OPHSD model through the plan--solve
  harness without the reference solution.}
  \label{tab:math-peak}
\end{table}

%-------------------------------------------------------------------
Following the same process, we first validate the effectiveness of the plan--solve harness. Shown in Table~\ref{tab:math-step0}, on the base model, plan--solve harness leads pass@1 by $17.83\%$ on average across the four benchmarks, giving OPHSD a strong teacher signal during training. Under this condition, we trace pass@8 over the training progress in Figure~\ref{fig:math-train} and report the best observed score of each method in Table~\ref{tab:math-peak}. OPHSD attains the highest average pass@8 (69.50\%), exceeding OPSD by 2.82\%, GRPO by 2.93\% and CRISP by 10.75\%. In contrast, OPSD achieved performance comparable to OPHSD during the early stages of training, but subsequently decreased rapidly. On OlympiadBench—which features a large volume of problems—OPHSD outperformed OPSD by 2.94\% and GRPO by 5.88\%, highlighting OPHSD's exceptional reasoning performance. We further test the trained OPHSD checkpoint through the plan--solve harness, which results 69.13\% on the same benchmarks, indicating that OPHSD has caught up with OPHSD + Harness and that the harness no longer provides a notable additional gain at inference time.

%-------------------------------------------------------------------
\subsection{What Does the Student Distilled from the Harness?}
\label{sec:internalize}
In this section, we analyze what the student learned from the harness. The harness we employed encompasses both retrieval-augmented reasoning based on interactions and plan-and-solve decomposition; therefore, we analyze these aspects separately.

\paragraph{Online text classification: retrieval-conditioned reasoning is written into the
weights.}

We use GPT-4o as an LLM-as-judge to detect whether the student's chain of thought generated without external retrieval spontaneously incorporates a case citation reasoning step (detailed in Appendix \ref{app:judge}).  Table~\ref{tab:text-cite} reports the rate at which the judge flags this behaviour at three stages (early / middle / late training). As shown, GRPO and OPSD never produce more than 10\% self-cited
references across training.  OPHSD reaches $\geq$ 75\% within the
first phase of training and stays $\geq$ 90\% at the end. This indicates the
trained student does not internalise the memory-bank contents but the shape of the harness's reasoning --- "here
is a similar case, it was decided as $X$ for reason $R$, so this case
is $X$" --- and re-emits that shape from the weights even when the
prompt contains no exemplar at all.

\paragraph{Mathematical Reasoning: the student inherits the harness-solvable problems
and preserves reasoning chain.}

\begin{figure}[t]

  \centering
  \begin{minipage}[t]{0.49\linewidth}
    \centering
    \includegraphics[width=\linewidth]{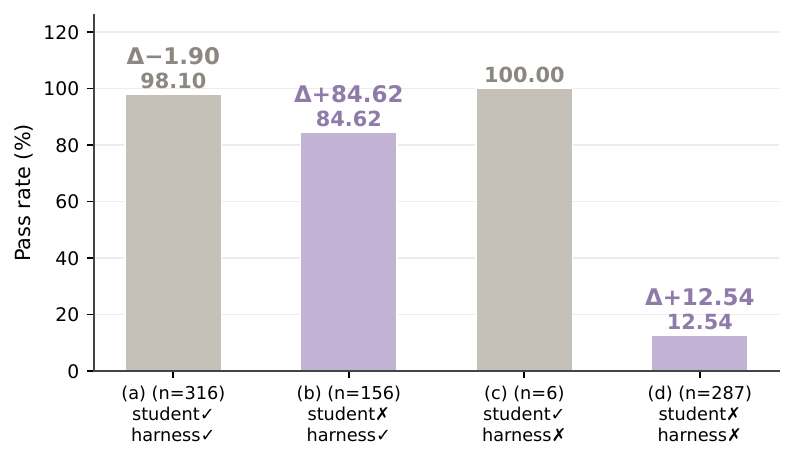}
    \caption{Relative changes in pass rate after OPHSD training. Mathematical reasoning problems are grouped by the capability gap between the base model (Qwen3-8B) and its plan--solve harness counterpart.}
    \label{fig:math-4bin}
  \end{minipage}\hfill
  \begin{minipage}[t]{0.49\linewidth}
    \centering
    \includegraphics[width=\linewidth]{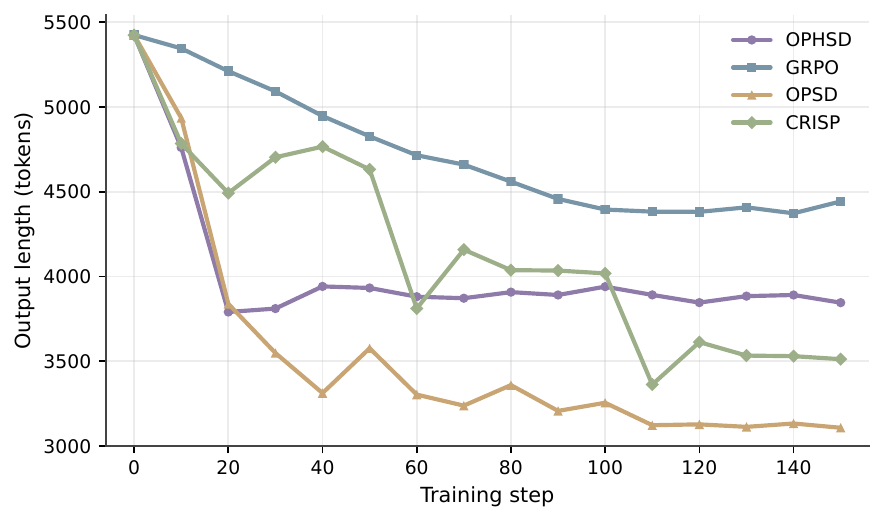}
    \caption{Average output length on math benchmarks. OPHSD ensures stability while reducing redundant reasoning.}
    \label{fig:math-len}
  \end{minipage}
\end{figure}

To identify the source of our distilled model's performance gains, we categorize the mathematical benchmark problems into four distinct groups. This categorization is based on the pass rate of the base model (Qwen3-8B) versus its harness-assisted counterpart. Figure~\ref{fig:math-4bin} illustrates the relative change in pass rate for each group following OPHSD training. We observe that, while strictly preserving the base model's original capabilities, OPHSD yields a 84.62\% relative improvement within Group b---the specific subset of problems that the base model could initially solve only when assisted by the harness. Furthermore, OPHSD enables the standalone model to successfully solve an additional 12.54\% of previously intractable problems (i.e., those initially unsolved in both settings). This is consistent with OPHSD successfully internalizing the external harness's reasoning power, delivering targeted and substantial procedural gains to the unassisted student.

Furthermore, we analyze the average response length on the
mathematical benchmarks without the harness in Figure~\ref{fig:math-len}. As can be observed, in comparison to GRPO, the output lengths of self-distillation-based methods (OPSD, OPHSD, CRISP) all decreased rapidly during the early stages of training. Under the supervision of a teacher, CRISP learned concise reasoning patterns and gradually declined as the teacher was iteratively updated. However, OPSD—in which the teacher directly provides the reference solution—led to a collapse in the student's reasoning length (after 40 steps); this outcome was subsequently validated by its performance decrease on the benchmarks (Figure \ref{fig:math-train}). In contrast, the distilled harness reasoning in OPHSD enabled the student to maintain stable performance on the test set while simultaneously reducing redundant reasoning.

%-------------------------------------------------------------------
\subsection{The Advantage Concentrates on the Hardest Problems}
\label{sec:lift}

\begin{figure}[t]
  \centering
  \includegraphics[width=\linewidth]{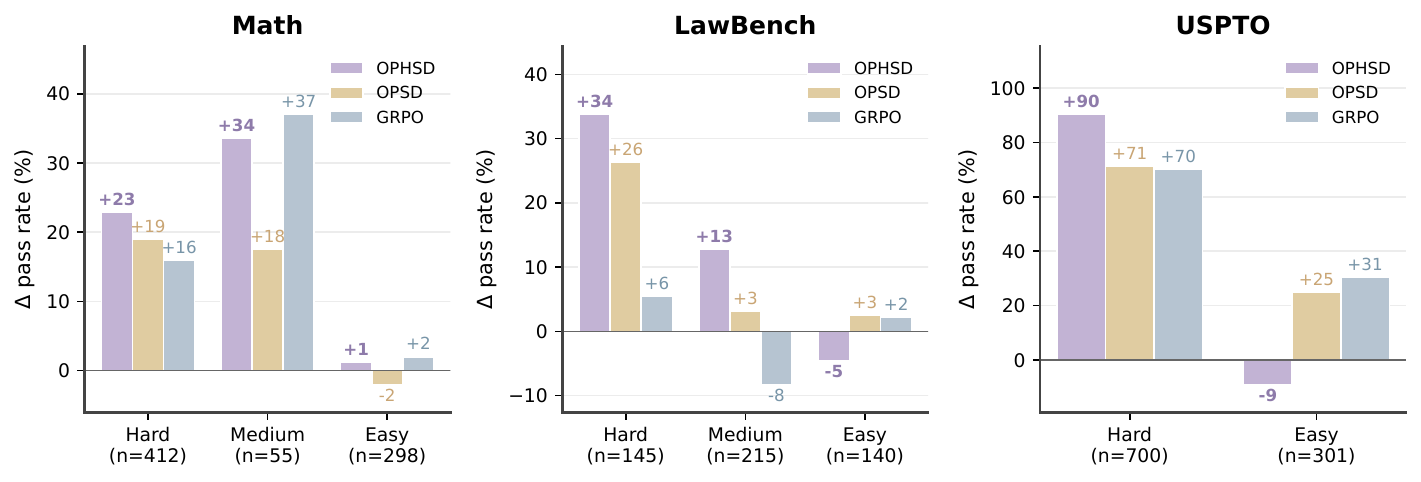}
  \caption{Performance gain across difficulty tiers. Samples are grouped into Hard, Medium, and Easy buckets based on the base model performance. For USPTO, the Medium bucket is omitted as its binary accuracy.} 
  \label{fig:lift}
\end{figure}

To check where OPHSD's advantage comes from we partition each
benchmark by per-question difficulty based on the base Qwen3-8B's per-question pass rate or F1 score into 3 groups: Hard $[0, 0.34)$, Medium $[0.34, 0.67)$, and Easy $[0.67, 1.0]$. For USPTO, since per-question outcomes are binary (correct/incorrect), the Medium bucket is empty by construction and thus omitted.
Figure~\ref{fig:lift} plots the gain after training per group. While easy groups are near ceiling and indistinguishable; the gap concentrates on the hard
groups, where OPHSD reaches +22.9\% on Math, +33.8\% on LawBench
and +90.4\% on USPTO --- all clearly above OPSD and GRPO.
This is consistent with OPHSD having successfully learned the enhanced reasoning capabilities provided by the harness. By internalizing the harness, it achieves a significant advantage over OPSD and GRPO on challenging problems.

%-------------------------------------------------------------------
% ============================================================================
% ============================================================================
\subsection{Case Study: Internalization and Harness Interference}
\label{sec:why}
% ============================================================================

Our experiments demonstrate that OPHSD-trained models achieve comparable, or even superior, performance at inference without the harness. On LawBench, the trained model already independently executes a nuanced, bidirectional ``recall-and-compare'' loop to evaluate candidate charges. Re-attaching the draft--verify harness overwrites this internal deliberation with a unidirectional external retrieval stream, where retrieved neighbors often cluster around obvious but incorrect charges. In this context, the external harness actively interferes with the student's learned reasoning (Appendix~\ref{app:case}). Similarly, on mathematical reasoning, the plan--solve harness forces explicit structural decomposition during training. The fully-distilled OPHSD student completely internalizes this rigorous planning habit, learning to actively flag combinatorial pitfalls. Consequently, while OPHSD achieves strong standalone performance, OPSD---trained merely to complete reference prefixes---lacks this structural rigor and routinely rushes into flawed computations (Appendix~\ref{app:math-case}).
In this sense, the harness acts as a ``scaffold'': essential during training to build internal capabilities, but redundant or even counterproductive once those capabilities are internalized. This also implies that certain components of the harness will be progressively removed as the model's intrinsic capabilities advance.
% ============================================================================
\section{Conclusion}
\label{sec:conclusion}
% ============================================================================

We introduced On-Policy Harness Self-Distillation (OPHSD), which internalizes the capabilities of harnesses directly into model parameters. By replacing static privileged information with dynamically orchestrated teacher signals, OPHSD trains the student to absorb complex reasoning strategies within the harness. Empirically, OPHSD achieves the highest standalone performance across text classification and mathematical reasoning tasks. The distilled model appears to internalize the harness's structural rigor, rendering external scaffolding redundant or even counterproductive at inference. Our results suggest that complex inference harnesses need not be permanent fixtures; they can serve as temporary training scaffolds whose procedural benefits are directly fed back into the base model, revealing an alternative path to enhancing model performance beyond the training data. Future work will explore distilling broader harness structures to internalize increasingly sophisticated workflows.

\paragraph{Limitations.}
Our empirical validation is currently scoped to two representative harnesses rather than a large-scale sweep across diverse harness designs. A detailed discussion regarding these boundaries and broader generalizability is provided in Appendix \ref{app:discussion}.

\clearpage

% --- 参考文献 ---
% 确保你有 reference.bib 文件
\bibliographystyle{plain}
\bibliography{reference}

\clearpage

% --- 附录 ---
\appendix
\appendix
\clearpage % 附录另起一页

% --- 自动生成附录目录的核心代码 ---
\section*{Appendix Content}
\startcontents[sections]
\printcontents[sections]{l}{1}{\setcounter{tocdepth}{2}}
\vspace{2em} % 目录和正文之间留点空隙
% ---------------------------------

\definecolor{boxQ}{HTML}{F4F1EC}     % warm grey, question
\definecolor{boxA}{HTML}{ECEEF3}     % cool grey, mode A
\definecolor{boxB}{HTML}{F1ECEC}     % rose grey, mode B
\definecolor{boxRule}{HTML}{8C8780}  % muted Morandi grey
\definecolor{boxPlan}{HTML}{E8F4F8}  % 浅灰蓝色：代表客观的 Plan/Harness
\definecolor{boxSolve}{HTML}{EDF7ED} % 浅茶绿色：代表正确的 OPHSD 解答
\definecolor{boxWrong}{HTML}{FDEDED} % 浅藕红色：代表错误的 OPSD 解答

\newtcolorbox{casebox}[2][]{%
  enhanced, breakable, sharp corners=south,
  colback=#2, colframe=boxRule, boxrule=0.4pt,
  left=8pt, right=8pt, top=6pt, bottom=6pt,
  fonttitle=\bfseries\small, coltitle=black,
  attach boxed title to top left={xshift=8pt, yshift=-7pt},
  boxed title style={colback=#2, colframe=boxRule,
                     sharp corners, boxrule=0.4pt},
  #1
}
\section{Implementation Details}
\label{app:impl}

\subsection{Harness}
\label{app:harness}

\paragraph{Draft--verify (text).}
We build the memory bank online using labelled training examples encountered prior to the current batch. To prevent train-test leakage during the harness baseline evaluation, we reset and re-populate the bank strictly from the test stream. We use the \texttt{BAAI/bge-small-zh-v1.5}~\cite{bge_embedding} model to compute sentence embeddings. For each harness call, we retrieve $k_d=5$ draft neighbours, $k_+=5$ confirmers, and $k_-=5$ challengers. We also implement a cold-start protection mechanism: until the memory bank accumulates at least 10 entries, we default to a direct single-forward pass. For generation, we set the temperature to $0.1$ and the maximum length to $8192$ tokens.

\paragraph{Plan--solve (math).}
We define the privileged input $z(x)=y^{\star}$ as the reference answer. During the plan phase, we use a temperature of $0.3$ to ensure structural stability, while in the solve phase, we increase the temperature to $0.6$. We set maximum length to $4096$ tokens for planner and $8192$ tokens for solver. At harness baseline evaluation, we remove $y^{\star}$, forcing the model to act as both planner and solver based solely on the problem statement.

\subsection{Baselines}
\label{app:baselines}

We train all methods on the Qwen3-8B base model using the verl~\citep{Sheng_2025} framework. We set the learning rate to $1 \times 10^{-6}$, the batch size to $64$, and the maximum generation length to $8192$ tokens. All experiments are conducted on 8$\times$H100 GPUs.

\textbf{GRPO}: We use a group size of $8$ and optimize the model with the standard GRPO policy-gradient objective. We set KL coefficient to 0.

\textbf{OPSD}: We apply reverse-KL distillation ($\mathrm{KL}(p_\text{teacher} \| p_\text{student})$). We set the teacher prefix to the verified solution, and fix the teacher policy to be the initial policy following the settings in the paper.

\textbf{CRISP}: Following the settings reported by \cite{sang2026crispcompressedreasoningiterative}, we optimize the same reverse-KL objective, and we use a static ``be concise'' system prompt without any ground-truth context. We sync the teacher weights from the student every $50$ steps (as the default setting in the paper).

\subsection{Evaluation}
\label{app:evaluations}
We train for 300 steps with evaluation every 15 steps for text classification, and 150 steps with evaluation every 10 steps for mathematical reasoning. A consistent evaluation protocol is adopted for the final results: for each method, we report the highest evaluation score obtained across the full training stage. (For mathematical reasoning, due to its inherent volatility, we report the average result of four runs.) This remains a fair comparison across different methods.

\subsection{LLM-as-Judge Protocol for the Cite-Rate Analysis}
\label{app:judge}
The cite-rate metric in Table~\ref{tab:text-cite} is computed by querying GPT-4o on each model's chain-of-thought produced \emph{without} the harness, and asking it to label two independent yes/no behaviours per trace: \emph{cites\_pseudo\_examples} (does the trace cite at least one specific past case as evidence?) and \emph{compares\_alternatives} (does it explicitly weigh at least two candidate labels?). We sample $N\!=\!40$ test instances per (scenario, method, training step); the cite-rate column in Table~\ref{tab:text-cite} aggregates the \emph{cites\_pseudo\_examples} field. The exact prompt is shown below. Prompts for all other components of OPHSD are released as part of our anonymous code repository linked in the abstract.

\begin{tcolorbox}[breakable, sharp corners=south, colback=boxQ, colframe=boxRule, boxrule=0.4pt, fonttitle=\bfseries\small, title={Judge prompt.}]
\small\ttfamily
SYSTEM: You are an expert annotator evaluating model reasoning traces on a domain-specific classification task (legal-charge prediction or chemical-reaction prediction).

\smallskip
USER: Below is a chain-of-thought produced by a model when answering one classification question. The model is asked to assign a label, and the trace explains its reasoning.

For this trace, decide TWO independent yes/no questions:

1. \textbf{cites\_pseudo\_examples} (bool): Does the trace explicitly recall or cite at least one specific past case / precedent / example as evidence to support its conclusion (e.g.\ ``in a similar case \dots'', ``previously, when \dots was charged with \dots'', ``consider example: \dots'')? General domain knowledge or definitional reasoning does NOT count.

2. \textbf{compares\_alternatives} (bool): Does the trace explicitly weigh at least two candidate labels against each other before committing to a final answer (e.g.\ ``this could be A or B, but \dots so it is A'')?

Then provide a one-sentence \textbf{reason} for your decision.

Return STRICT JSON with exactly the keys: \{"cites\_pseudo\_examples": <bool>, "compares\_alternatives": <bool>, "reason": <string>\}.

\smallskip
Trace: \{trace\}\\
Final label: \{label\}
\end{tcolorbox}

% ============================================================================
\section{Case Study}
\subsection{Text Classification: Where the Harness Narrows the Trained Student's Deliberation}
\label{app:case}
% ============================================================================

We analyzed two modes of reasoning.
\textbf{Mode~A} is a single forward pass without any harness of the trained student.
\textbf{Mode~B} is the same model within the draft--verify harness.
We take one question (LawBench-3-3, idx~44) for end-to-end
inspection, in which Mode A surpasses Mode B. The data is translated from Chinese.

% ── The question ─────────────────────────────────────────────────────────────
\subsubsection*{The Question}

\begin{casebox}[title={Question (LawBench-3-3, idx~44)}]{boxQ}
\small
At around 16:30 on 7 February 2015 the defendants Wang and Gu were
drinking at a restaurant in Tumen City.  After a verbal dispute with
the victim Liu over a trivial matter, the three of them engaged in a
mutual fight.  At 18:09 the same day, while being interviewed by the
police, Liu felt unwell and was taken to a hospital, where he died at
20:25 despite resuscitation.  \emph{The forensic report concludes that
Liu died from an acute cardiac event triggered by the verbal and
physical altercation, on a background of coronary heart disease, and
that there is a causal relationship between the trauma and the death.}
Liu sustained bilateral comminuted nasal-bone fractures, rated as a
class-II minor injury.

\smallskip
\textbf{Gold label:}~\texttt{negligent homicide}.
\end{casebox}

\noindent
This is a textbook borderline question.  Death follows from a fight,
\emph{but} the proximate cause is an acute cardiac event triggered by
the trauma rather than the trauma itself.  The legally correct charge
is the negligent offence, not the obvious \texttt{intentional injury}
one would infer from a routine fight pattern.

% ── Mode A ───────────────────────────────────────────────────────────────────
\subsubsection*{Mode A --- Single Forward, No Harness}

The trained student's chain of thought stages a \emph{bidirectional}
deliberation between the two candidate charges:

\begin{casebox}[title={Mode A reasoning excerpt (translated)}]{boxA}
\small\itshape
\ldots The prosecution argues that Wang and Gu's conduct constitutes
\texttt{intentional injury}\ldots\ \textbf{but I should also consider
other possible charges, such as \texttt{negligent homicide} or
\texttt{intentional injury causing death}.}\ldots

\smallskip
\ldots If their conduct amounted to negligence in a fight --- for
example, having no foreknowledge that Liu suffered from coronary heart
disease --- \textbf{then their offence may instead constitute
\texttt{negligent homicide}.}  However, in routine fight-pattern cases,
injuries inflicted in an affray are typically classified as
\texttt{intentional injury}\ldots
\end{casebox}

\noindent\textbf{Mode A verdict:}~\texttt{[charge] intentional injury;
negligent homicide \textlangle eoa\textrangle}
(set-F1~$= 0.67$; \emph{contains the gold label}).

% ── Mode B ───────────────────────────────────────────────────────────────────
\subsubsection*{Mode B --- Trained Model with the Harness}

\paragraph{Draft phase.}
The five memory-bank neighbours retrieved by cosine similarity all
carry \texttt{intentional injury}:

\begin{casebox}[title={Top-5 retrieved neighbours (draft prompt)}]{boxB}
\small
\begin{tabular}{@{}rl@{}}
  $0.814$ & \texttt{intentional injury; unlawful detention}\\
  $0.811$ & \texttt{unlawful possession of firearms; intentional injury; affray}\\
  $0.808$ & \texttt{intentional injury; obstruction of justice}\\
  $0.799$ & \texttt{intentional injury; harbouring offenders}\\
  $0.797$ & \texttt{intentional injury}
\end{tabular}

\smallskip\noindent
None of them carries \texttt{negligent homicide}.
\end{casebox}

Conditioned on these one-sided exemples, the model's draft chain of
thought collapses the previously two-way weighing into a one-way
confirmation:

\begin{casebox}[title={Mode B draft reasoning excerpt (translated)}]{boxB}
\small\itshape
\ldots In similar cases, when injurious conduct directly causes another
person's death, \textbf{the offence is typically classified as
\texttt{intentional injury}}\ldots\ In the first exemplar, the
defendant struck the victim and inflicted serious injuries ---
classified as \texttt{intentional injury}; in another exemplar, an
affray caused another person's death --- \textbf{also classified as
\texttt{intentional injury}}\ldots
\end{casebox}

\noindent\textbf{Draft verdict:}~\texttt{[charge] intentional injury
\textlangle eoa\textrangle}.
\texttt{Negligent homicide} is no longer even \emph{mentioned} in the trace.

\paragraph{Verify phase.}
The challenger neighbours (cases whose label set differs from the
draft) come from entirely unrelated semantic clusters:

\begin{casebox}[title={Top-5 challenger neighbours (verify prompt)}]{boxB}
\small
\begin{tabular}{@{}rl@{}}
  $0.750$ & \texttt{unlawful trespass}\\
  $0.730$ & \texttt{rape}\\
  $0.718$ & \texttt{murder}\\
  $0.698$ & \texttt{theft; operating a gambling house}\\
  $0.693$ & \texttt{theft; drug trafficking}
\end{tabular}

\smallskip\noindent
None carries \texttt{negligent homicide}; all are semantically
unrelated to the fight-pattern question facts.
\end{casebox}

Asked to verify or correct the draft, the model has nothing in front
of it that points back at \texttt{negligent homicide}.

\noindent\textbf{Mode B verdict:}~\texttt{[charge] intentional injury
\textlangle eoa\textrangle} (set-F1~$= 0$).

% ── Analysis ─────────────────────────────────────────────────────────────────
\subsubsection*{Analysis}

Three things happen on the same OPHSD-trained student.

\begin{enumerate}
  \item \textbf{Mode A's internal retrieval is bidirectional.}
    The student's chain of thought self-stages a
    \emph{recall $\rightarrow$ compare $\rightarrow$ decide} loop that
    \emph{retains} \texttt{negligent homicide} as a viable candidate
    alongside \texttt{intentional injury}.

  \item \textbf{Mode B's external retrieval is unidirectional.}
    Cosine nearest neighbours in the training memory bank are dominated
    by routine \texttt{intentional injury} fight cases; the challenger
    slot, meant to expose the student to contrastive alternatives, gets
    filled with semantically unrelated offences because no
    \texttt{negligent homicide} neighbour is close enough to the
    fight-pattern question facts.  The harness amplifies exactly the
    one-sided evidence the student had been weighing \emph{against}
    internally.

  \item \textbf{Harness interfered with the trained student model.}
    Asked to verify a draft already conditioned on five
    \texttt{intentional injury} exemplars, the student abandons the
    bidirectional weighing it ran in Mode~A and locks onto the
    single-stream harness evidence.  The internally tracked
    \texttt{negligent homicide} candidate is pushed out of the answer
    set, and the gold label disappears.
\end{enumerate}

% ============================================================================
\subsection{Mathmatical Reasoning: Plan-Structured Supervision Improves Reasoning Precision}
\label{app:math-case}
% ============================================================================

To illustrate how plan-solve supervision shapes the trained model's
reasoning, we trace a single OlympiadBench question (idx~649) through the
full pipeline: the plan the harness teacher generated, the solution the
OPHSD-trained student produced at inference, and the mistake the
OPSD-trained student made.

% ── The question ─────────────────────────────────────────────────────────────
\subsubsection*{The Question}

\begin{casebox}[title={Question (OlympiadBench, idx~649)}]{boxQ}
\small
Let $f(x)=x^{1}+x^{2}+x^{4}+x^{8}+x^{16}+x^{32}+\cdots$.
Compute the coefficient of $x^{10}$ in $f(f(x))$.

\smallskip
\textbf{Answer:}~$40$.
\end{casebox}

\noindent
The composition $f(f(x))=\sum_{k\ge0}[f(x)]^{2^{k}}$ reduces to
summing the number of \emph{ordered} $2^k$-tuples of powers of two that
add to 10.  A common pitfall is treating $(2,8)$ and $(8,2)$ as a single
\emph{unordered} pair when computing the $k=1$ contribution, which yields
1 instead of the correct 2, and propagates to a final answer of 39.

% ── Harness plan ─────────────────────────────────────────────────────────────
\subsubsection*{Step 1 (Training): the Harness Generates a Plan}

During training, the plan-solve harness wraps the student model.  In the
plan phase the model is asked to outline a solution \emph{without}
solving it.  On this question, the plan explicitly flags the distinction
that matters:

\begin{casebox}[title={Harness teacher --- plan phase (excerpt)}]{boxPlan}
\small
\ldots For each $k$, compute the coefficient of $x^{10}$ in $[f(x)]^{2^k}$
\ldots

\textbf{For $k=1$:} $[f(x)]^2$, find all pairs of powers of two that
sum to 10:
\begin{itemize}
  \item $2+8=10$, and \textbf{$(2,8)$ and $(8,2)$ are distinct ordered
    pairs}
  \item Total: \textbf{2} ways
\end{itemize}
\textbf{For $k=2$:} $[f(x)]^4$: combinations $(2,2,2,4)$ give 4
permutations; $(1,1,4,4)$ give 6 permutations. Total: \textbf{10} ways.\\
\textbf{For $k=3$:} $[f(x)]^8$: $(1^6,2^2)$ gives $\binom{8}{2}=28$.
Total: \textbf{28} ways.
\end{casebox}

\noindent
The solve phase follows the plan and produces $0+2+10+28=\boxed{40}$.
Over eight stochastic harness rollouts at training step 80, four out of
eight harness traces are correct; the student receives correct plan-based
supervision on this problem.

% ── OPHSD student ────────────────────────────────────────────────────────────
\subsection*{Step 2 (Inference): the OPHSD Student Catches the Pitfall}

At inference the trained OPHSD student runs without any harness.  Its
chain of thought reproduces the plan-shaped analysis and explicitly
self-corrects on the ordered-pair question:

\begin{casebox}[title={OPHSD student reasoning excerpt (OlympiadBench idx~649)}]{boxSolve}
\small\itshape
\ldots\ For $[f(x)]^2$, possible pairs where $a+b=10$, $a,b\in\{1,2,4,8,\ldots\}$:
$(2,8)$, $(8,2)$\ldots

\medskip
\textbf{Wait,} but since multiplication is commutative, $(2,8)$ and $(8,2)$
might seem the same.  \textbf{But in generating functions, the coefficient
of $x^{10}$ is the number of \emph{ordered} pairs $(a,b)$ with $a+b=10$.
So $(2,8)$ and $(8,2)$ are two distinct ordered pairs.  Therefore, the
coefficient is~\textbf{2}.}

\ldots\ Total: $0+2+10+28=\boxed{40}$.
\end{casebox}

\noindent
OPHSD scores 7/8 correct on this question.

% ── OPSD student ─────────────────────────────────────────────────────────────
\subsubsection*{Conversely, the OPSD Student Makes the Error}

The OPSD student, trained on ground-truth-prefixed traces that do not
carry explicit enumeration plans, never encounters the ordered-vs-unordered
distinction during training.  Its chain of thought conflates the two:

\begin{casebox}[title={OPSD student reasoning excerpt (OlympiadBench idx~649)}]{boxWrong}
\small\itshape
\ldots For $[f(x)]^2$, we look for pairs of exponents $a,b\in\{1,2,4,8\}$
such that $a+b=10$.
\textbf{The only valid pair is $(2,8)$, which occurs once.
So the coefficient from $[f(x)]^2$ is~\textbf{1}.}

\ldots\ Total: $0+1+10+28=\boxed{39}$.
\end{casebox}

\noindent
OPSD scores 0/8 correct on this question.

% ── Analysis ─────────────────────────────────────────────────────────────────
\subsubsection*{Analysis}
The divergence stems from structural rigor rather than reasoning capacity. While both models approach the problem identically, OPSD fails at a critical conceptual step: distinguishing ordered from unordered sequences.

Plan--solve supervision mitigates this by design. During training, the harness forces the model to articulate potential pitfalls before any arithmetic occurs. Consequently, the OPHSD student internalizes the necessity of formalizing enumeration semantics prior to calculation. In contrast, OPSD---trained merely to complete reference prefixes---receives no such structural grounding. It routinely bypasses the crucial setup phase and rushes into flawed computations.

This example demonstrates how programmatic harness supervision embeds rigorous analytical patterns, shaping the model's deliberation far more deeply than standard ground-truth imitation.

\section{Discussion: Which Harnesses Are OPHSD-Compatible?}
\label{app:discussion}

The boundary established in \S\ref{sec:method_ophsd}---that procedural reasoning priors are internalizable while real-time external grounding is not---admits a principled and finer-grained taxonomy. Because the unassisted student observes only the input $x$ at inference, a harness is OPHSD-compatible precisely to the extent that its performance gains can be reproduced exclusively from $x$ and the model's pre-existing parametric memory. We categorize this compatibility into three distinct regimes.

\paragraph{Procedural priors over $x$ (Fully Distillable).}
Here, the harness restructures how the model processes $x$ without injecting novel information that the model could not theoretically derive on its own. Techniques such as plan-and-solve, multi-step decomposition, self-verification, and structured chain-of-thought templates fall into this category. When a privileged reference solution $z(x)$ is utilized during training, it acts purely as a \emph{strategy demonstrator}: it teaches the student the mechanics of decomposition without being required at inference. The plan--solve harness evaluated in this paper exemplifies this class, allowing OPHSD to recover essentially the entirety of the harness-induced performance gain.

\paragraph{Procedural shape over dynamic content (Partially Distillable).}
In this regime, the harness relies on context available exclusively during training---such as an external memory bank, asymmetric knowledge indices, or a stream of labeled examples. While the structural \emph{shape} of the reasoning procedure (e.g., ``draft a candidate, compare with similar precedents, accept or revise'') is highly transferable, the specific queried content cannot be internalized. Our draft--verify harness operates within this space: although the external memory bank is removed at inference, the distilled student spontaneously executes the case-based comparison pattern using only its internal parametric knowledge (evidenced by the $>$\,90\% cite rate in \S\ref{sec:internalize}). The extent to which OPHSD recovers the full harness gain depends heavily on whether the model's parametric prior contains sufficient domain knowledge to populate this internalized reasoning shape; when it does not, the procedural shape still triggers, but operates on weaker evidence.

\paragraph{Real-time external interaction (Non-Distillable).}
This regime involves harnesses that consult oracles whose outputs are fundamentally underivable from $x$ or the model's weights: live web search for current events, code execution for precise numerical simulation, or interaction with a stateful environment. Because the harness's advantage is strictly informational rather than procedural, OPHSD cannot internalize the execution itself. While it is theoretically plausible that OPHSD could distill the procedural scaffolding surrounding the tool call---such as learning when to invoke the tool or how to format the query---empirically validating this capability lies beyond the scope of this paper and remains a promising direction for future research.

\paragraph{Practical Heuristic for OPHSD.}
Before applying OPHSD to a novel task, a reliable diagnostic is to perform an inference-time ablation of the privileged context $z(x)$ from the harness. If a substantial portion of the performance gain over the base model is preserved, the harness operates primarily via procedural priors, making it highly compatible with OPHSD. Conversely, if performance collapses, the harness relies heavily on real-time external grounding (Regime 3). More broadly, the larger the proportion of a harness's benefit that derives from how it structures reasoning rather than what dynamic information it retrieves, the greater the share of procedural power OPHSD will successfully transfer back into the base model.

\section{Broader Impacts}
\label{app:broader_impacts}
By internalizing inference-time harnesses, OPHSD significantly reduces deployment latency, token costs, and engineering complexity. However, this standalone capability also amplifies potential misuse risks, as models can now execute advanced reasoning without the safety nets typically integrated into external workflows. Consequently, internalized procedural reasoning should not replace ground-truth verification, and standard safeguards—such as human-in-the-loop review—remain strictly necessary for high-stakes domains.

\end{document}